\documentclass[11pt,a4paper]{article}
\usepackage{eacl2021}
\usepackage{times}
\usepackage{latexsym}

\usepackage[htt]{hyphenat}

\usepackage{microtype}

\aclfinalcopy
\usepackage{graphicx}
\usepackage{tabularx}
\usepackage{soul}
\usepackage{tikz}

\usepackage[utf8]{inputenc}

\usepackage{color}

\usepackage[tbtags]{amsmath}
\usepackage{booktabs}
\usepackage{multirow}
\usepackage{xcolor}
\usepackage{colortbl}
\usepackage{standalone}
\usepackage{pgfplots}
\pgfplotsset{compat=newest}
\usepgfplotslibrary{colorbrewer}
\usepgfplotslibrary[groupplots,dateplot]
\usetikzlibrary[patterns,shapes.arrows]
\unexpanded{}
\unexpanded{}

\definecolor{PaleYellow}{RGB}{255,230,204}
\definecolor{PaleRed}{RGB}{248,206,204}
\definecolor{PalePurple}{RGB}{225,213,231}
\definecolor{AquaBlue}{RGB}{176,227,230}

\usepackage{graphicx}                                               %
\usepackage{hyperref}        
\usepackage{xurl}
\hypersetup{breaklinks=true}
\usepackage[hyphenbreaks]{breakurl}
\usepackage{booktabs}                                               %
\usepackage{multirow}                                               %
\usepackage{nicefrac}                                               %

\usepackage{tablefootnote}      
\usepackage{float}

\title{Native Language Identification with Big Bird Embeddings}

         \author{Sergey Kramp, Giovanni Cassani, Chris Emmery \\
  CSAI, Tilburg University \\ 
  \texttt{sergey.kramp@gmail.com }
  }

\date{}
\begin{document}
\maketitle

\begin{abstract}
Native Language Identification (NLI) intends to classify an author's native language based on their writing in another language.
Historically, the task has heavily relied on time-consuming linguistic feature engineering, and
transformer-based NLI models 
have thus far failed to offer effective, practical alternatives. 
The current work investigates if input size is a limiting factor, and 
shows that classifiers trained using Big Bird embeddings outperform linguistic feature engineering models by a large margin %
on the Reddit-L2 dataset. Additionally, we provide further insight into input length dependencies, show consistent out-of-sample performance, and qualitatively analyze the embedding space. 
Given the effectiveness and computational efficiency of this %
method, we believe it offers a promising avenue for future NLI work.
\end{abstract}

\section{Introduction} \label{sec:introduction}

Native Language Identification (NLI) operates under the assumption that an author's first language (L1) produces discoverable patterns in a second language (L2) \cite{language-transfer-odlin,influence-of-l1-on-l2}. Classifying one's native language proves highly useful in various applications, such as in language teaching, where customized feedback could be provided based on the learner native language; in fraud detection, where identifying an unknown author's native language can aid in detecting plagiarism and web fraud; and in consumer analytics. NLI models historically relied on handcrafted linguistic patterns as input features \cite{koppelOriginalNLI,improving-nli-with-spelling-errors,tetreault-shared-task-1,cimino2013-dependency-parsing}; however, such representations are unlikely to capture all required nuances and complexities of this task \cite{moschitti2004-linguistic-features-limitations}, in particular on noisier sources of data.

Current transformer models \cite{DBLP:conf/nips/VaswaniSPUJGKP17} have shown success in such challenges \cite{gpt3-paper} %
but are often limited by input size. %
 This is particularly problematic %
for NLI %
which often deals with long texts, such as essays, documents or social media posts. %
Our work is the first to employ long-form transformer models to overcome these limitations. We train a simple logistic regression classifier using embeddings from a fine-tuned Big Bird model, %
and demonstrate it significantly outperforms a similar classifier trained using costly handcrafted feature representations.\footnote{Code and models available at  \url{github.com/SergeyKramp/mthesis-bigbird-embeddings}.} %

\section{Related Work}

Seminal NLI work by \newcite{koppelOriginalNLI} used function words, character $n$-grams, and handcrafted error types as features---restricted to 1000 articles in 5 languages.
The TOEFL-11 dataset \cite{toefl11} proved a fruitful resource for two NLI shared tasks \cite{tetreault-shared-task-1,malmasi2017-shared-task-2}. %
However, its controlled collection environment and limited range of topics affected generalization of traditional linguistic features to noisy Internet data \cite{noisy-social-media}.
An example of such noisy data is the Reddit-L2 dataset \cite{rabinovich-2018-reddit-l2}; the current de facto benchmark for NLI, which we employ in the current study as well. %

Despite various attempts using neural architectures \cite{ircing-2017-shared-task2-NN,bjerva-etal-2017-convolutional,franco-salvador-etal-2017-bridging}, the current best performance on the Reddit-L2 dataset was obtained by \newcite{goldin-etal-2018-native} using a logistic regression classifier trained on a combination of linguistic features. We will implement (and thereby directly compare to) their work in our experiments.

Most related are two studies using transformers for NLI.  \newcite{steinbakken-gamback-2020-bert-nli} fine-tuned BERT on a less challenging part of the Reddit-L2 dataset \cite{bert-original-paper-2018}; stand-alone, and in an ensemble of classifiers. 
 \newcite{lotfi-etal-2020-gpt2} fine-tuned GPT-2 \cite{Radford2019LanguageMA} per language in the TOEFL-11 dataset (i.e., 11 in total), using the lowest loss among them to classify an instance. %
Our method offers a stand-alone transformer model approach with a much lower computational footprint. We will evaluate performance on the Reddit-L2 split with little to no information related to (linguistic) geography.

\newcite{rabinovich-2018-reddit-l2} have used hierarchical clustering to investigate the relationship between an author's native language and their lexical choice in English. %
Using word frequency and word embeddings of English words, they measured distances between 31 L1s. Languages from the same family appear closest in this space. They further suggested that authors with a similar L1 have similar idiosyncrasies in their English writing. Hence, given an accurate model, we expect to find similar representations in our embedding spaces.

\section{Methodology}

We test if Big Bird embeddings are a suitable application of the transformer architecture for NLI, thereby mostly replicating the experimental design of \newcite{goldin-etal-2018-native}.

\subsection{Data} \label{sec:dataset}

We used a derivative of the Reddit-L2 dataset, first introduced as L2-Reddit by \newcite{rabinovich-2018-reddit-l2}, and used in \newcite{goldin-etal-2018-native}. The raw data\footnote{Via: \url{http://cl.haifa.ac.il/projects/L2/}} consists of $200\text{M}$ sentences ($\sim 3\text{B}$ tokens), and spans the years 2005-2017 and used the old (i.e., free) Reddit API. Data collection used flairs that report country of origin on subreddits discussing European politics, yielding a total of $45K$ labeled native and non-native English-speaking users and their entire post history. Between-group language proficiency was accounted for through several syntactic and lexical metrics, and languages with fewer than 100 authors were removed. Each author profile was split per 100 sentences, and these ``chunks'' were subsequently divided in two splits: one partition with subreddits discussing European politics (referred to as the \texttt{europe} partition), and a second partion from all other subreddits (the \texttt{non\textunderscore europe} partition). 

\paragraph{Sampling}

For L1 identification, we regrouped the Reddit-L2 dataset on native language rather than nationality. After filtering predominantly multi-lingual countries, this resulted in 23 labels. %
We found that the majority are native English speakers,  with Dutch native speakers constituting the second largest part, and that there is a stronger label imbalance in the \texttt{non\textunderscore europe} partition than in the \texttt{europe} partition. 

In accordance with \newcite{goldin-etal-2018-native}, the data was balanced through downsampling by randomly selecting 273 and 104 authors respectively for each language in our two partitions. These author proportions are based on the least represented language in each partition: Slovenian and Lithuanian.
Similarly, to reduce the skew that highly active authors for a given language add to the data, the amount of chunks per author was capped. These were randomly sampled until the median value per author; 17 for the \texttt{non\textunderscore europe} partition, and 3 for the \texttt{europe} one.

\paragraph{Preprocessing} \label{sec:preprocessing}

For this, we removed redundant blank spaces and replaced all URLs with a special token. While minimal, these changes improved classification performance across the board.

\paragraph{Splitting} \label{sec:data-splitting}

We split the \texttt{non-europe} partition on chunk level\footnote{Splitting by authors had negligible effects.} into equal fine-tuning ($D_\text{tune}$), and training and testing ($D_\text{exp}$) parts. We hypothesized that due to the size and variety of the \texttt{non\textunderscore europe} partition, it is a more realistic, challenging part of the data. Unlike the \texttt{europe} partition used by \newcite{steinbakken-gamback-2020-bert-nli}, it covers a variety of topics and contains fewer context words (e.g., countries and nationalities) that might pollute classification. Instead, we dedicated the entire \texttt{europe} partition to conduct an out-of-sample evaluation. We refer to this data as $D_\text{oos}$. As this part of the data contains texts on topics not seen in $D_\text{tune}$ and $D_\text{exp}$, this allows us to gauge the context specificity of our representations.

\subsection{Feature Engineering Baseline} \label{sec:feature-engineering}
Here we describe how the linguistic features\footnote{For comparison sake, we did not optimize these further.} (5186 total) were constructed. We followed \newcite{goldin-etal-2018-native} or found equivalents to the features used in their work. These were extracted for each chunk.

\paragraph{$n$-Grams} To create word unigram and character trigram features, we used \texttt{scikit-learn} \cite{scikit-learn}. Both vectorizers were fit on the text chunks of $D_\text{exp}$ and cut off at the 1000 most common $n$-grams.

\paragraph{Edit Distance and Substitution} To collect the spelling errors in the data, we used the \texttt{symspellpy}\footnote{\href{https://github.com/mammothb/symspellpy}{\texttt{github.com/mammothb/symspellpy}}} package. For each misspelled word in $D_\text{exp}$, we obtained its closest correction with a maximal edit distance of 2. Words for which no correction was found were ignored. Next, we tracked which characters were inserted, deleted or replaced to arrive at the correction. This resulted in a substitution frequency list, of which the top 400 were used. The number of occurrences of each substitution type in the chunk was used as features. Subsequently, for each chunk we aggregated the Levenshtein distance between all words and their corrections, and divided this by the total number of words, giving the average edit distance.

\paragraph{Grammar, POS, Function Words, Length}
For the other features, each chunk in $D_\text{exp}$ was split into individual sentences (by \texttt{\textbackslash n}). The grammar error features were extracted using the \texttt{LanguageTool} Python wrapper\footnote{\href{https://github.com/jxmorris12/language_tool_python}{\texttt{github.com/jxmorris12/language\_tool\_py}}} to produce a list of errors for all sentences in $D_\text{exp}$. In total, we found 2017 error types in the data and used all of them as binary features (i.e., the presence or absence of a grammar error in that chunk). POS trigrams were created through \texttt{nltk}\footnote{We used the pre-trained Averaged Perceptron Tagger in combination with the Punkt Tokenizer.} \cite{nltk}, and their top 300 used as features. For the function word frequency features, we used a list of 467 function words taken from \newcite{Volansky2015OnTF}. For average sentence length, we removed all non-alphanumeric symbols of length 1 to exclude punctuation and special symbols, then divided sentence length (on word level) by the total number of sentences in a chunk (i.e., 100). %

\subsection{Transformer Model} \label{sec:transformer-models}
The main focus of this study was to find an efficient method to use transformers for NLI. To this end, we chose Big Bird (\texttt{google/bigbird-roberta-base}) from the Hugging Face Model Hub \cite{transformers}, as it provides a relatively large context length of 4096 tokens while fitting on one GPU.\footnote{We used an Nvidia Titan X with 12 GB of VRAM.}

\paragraph{Fine-tuning}
We fine-tuned all layers of Big Bird on $D_\text{tune}$ using the hyperparameters specified in the original paper: Adam \cite{DBLP:journals/corr/KingmaB14} to optimize with the learning rate set to $10^{-5}$ and epsilon to $10^{-8}$. Warm-up on 10\% of all training inputs ran during the first epoch. Fine-tuning ran for 3 epochs totaling 15 hours. 
Due to memory constraints, we used an input size of 2048, with a batch size of 2. Chunks that were shorter were padded to match the input length; longer inputs were split into sub-chunks (padded to full length). 

\paragraph{Embedding Representation}

In order to compare Big Bird to lingustic features, we do not train Big Bird end-to-end. Rather, we extract its embeddings (either pre-trained from the Model Hub or our own fine-tuned version) and use them as input for a downstream classifer. For tokenization, we used the matching pre-trained tokenizers from \texttt{transformers},\footnote{\href{https://github.com/huggingface/transformers}{\texttt{github.com/huggingface/transformers}}} which were fine-tuned during our experiments. We added \texttt{[CLS]} at the beginning of the first sentence of each chunk, and manually inserted a separator token between each sentence in the chunk and at the end of the chunk.

Following \newcite{bert-original-paper-2018}, we used the last hidden states for \texttt{[CLS]} as 768-dimensional embedding features per chunk. We experimented with 3 token input sizes: 512 (BERT's input size), 2048 (size also used when fine-tuning), and 4096  (Big Bird's maximum input size). %

\begin{table}[t]
    \centering
    \footnotesize
    \begin{tabular}{lrrr}
        \toprule
        \textsc{model} & \textsc{dur} & \textsc{acva} & \textsc{oosa} \\
        \midrule
        $Feature Engineering$   & 13.00 & .475      & .637 \\
        $BigBird_{512}$         &  0.27 & .364      &    - \\
        $BigBird_{512\_tuned}$  &  0.27 & .432      &    - \\
        $BigBird_{2048}$        &  2.50 & .493      & .774 \\
        $BigBird_{2048\_tuned}$ &  2.50 & \bf{.654} & \bf{.855} \\      
        $BigBird_{4096}$        &  3.00 & .500      &    - \\
        $BigBird_{4096\_tuned}$ &  3.00 & .635      &    - \\
        \bottomrule
    \end{tabular}
    \caption{The models (\textsc{name}) annotated with their input dimensions and if they were fine-tuned, how long feature extraction took on $D_\text{exp}$ (\textsc{dur}, in hours), their average cross-validation accuracy scores on $D_\text{exp}$ (\textsc{acva}) and accuracy scores on $D_\text{oos}$ (\textsc{oosa}).}
    \label{tab:results}
\end{table}

\section{Experimental Setup} \label{sec:experiment-design}
For our main experiment, we followed the experimental design in \newcite{goldin-etal-2018-native}:

\subsection{Main Experiment} \label{sec:classifiers}
We trained a logistic regression classifier 
on the output of each feature extractor. To further establish an equal ground for comparison, we did not tune the hyperparameters of these classifiers. 
Hence, we adopted \texttt{scikit-learn}'s default parameters: $\ell_2$ normalization, $C = 1$, L-BFGS \cite{DBLP:journals/mp/LiuN89} for optimization, and maximum  iterations set to 1000. To gauge the robustness of each classifier's performance, we used 10-fold cross-validation (CV); in particular, we looked at the average CV accuracy score of each classifier. Given that we adhere to prior work and accordingly balanced the labels, we found that additional metrics provided little added insight.

\subsection{Embedding Space Analysis} \label{sec:method-embeddings-space-analysis}
Following \newcite{rabinovich-2018-reddit-l2}, we used hierarchical clustering to analyze how each native language is represented in the 768-dimensional embedding space. %
We used the best performing pre-trained and fine-tuned Big Bird models from our main experiment to compute %
the centroids (23 in total) on $D_\text{exp}$. Subsequently, we used \texttt{scipy}'s \cite{2020SciPy-NMeth} implementation of Ward's linkage function \cite{ward1963hierarchical} to create a cluster dendrogram, and \texttt{scikit-learn}'s default implementation of Principal Component Analysis \cite[PCA]{hotelling1933analysis,tipping1999probabilistic} to visualize the centroids in a 2-dimensional space.

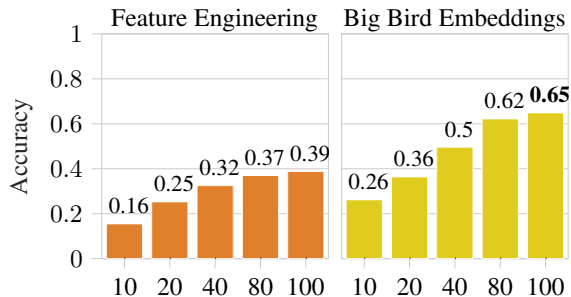
\begin{figure}[t]
    \centering
    \definecolor{darkslategray38}{RGB}{38,38,38}
\definecolor{darkslategray66}{RGB}{66,66,66}
\definecolor{goldenrod22320431}{RGB}{223,204,31}
\definecolor{lightgray204}{RGB}{204,204,204}
\definecolor{peru22412844}{RGB}{224,128,44}

\begin{tikzpicture}

\begin{groupplot}[group style={group size=2 by 1, horizontal sep=0.2cm}]
\nextgroupplot[
axis line style={lightgray204},
enlargelimits=false,
height=5cm, %
tick align=outside,
title style={yshift=-1.8ex},
title={Feature Engineering},
unbounded coords=jump,
width=5cm, %
x grid style={lightgray204},
xlabel=\textcolor{darkslategray38}{},
xmajorticks=true,
xtick pos=bottom,
xmin=-0.5, xmax=4.5,
xtick style={color=lightgray204},
xtick={0,1,2,3,4},
xticklabels={10,20,40,80,100},
y grid style={lightgray204},
ylabel=\textcolor{darkslategray38}{Accuracy},
ymajorgrids,
ymajorticks=true,
ytick pos=left,
ymin=0, ymax=1,
ytick style={color=lightgray204}
]
\draw[draw=white,fill=peru22412844] (axis cs:-0.4,0) rectangle (axis cs:0.4,0.155);
\draw[draw=white,fill=peru22412844] (axis cs:0.6,0) rectangle (axis cs:1.4,0.253);
\draw[draw=white,fill=peru22412844] (axis cs:1.6,0) rectangle (axis cs:2.4,0.325);
\draw[draw=white,fill=peru22412844] (axis cs:2.6,0) rectangle (axis cs:3.4,0.37);
\draw[draw=white,fill=peru22412844] (axis cs:3.6,0) rectangle (axis cs:4.4,0.388);
\addplot [line width=1.08pt, darkslategray66]
table {%
0 0
0 0
};
\addplot [line width=1.08pt, darkslategray66]
table {%
1 0
1 0
};
\addplot [line width=1.08pt, darkslategray66]
table {%
2 0
2 0
};
\addplot [line width=1.08pt, darkslategray66]
table {%
3 0
3 0
};
\addplot [line width=1.08pt, darkslategray66]
table {%
4 0
4 0
};
\draw (axis cs:-0.2,0.165) node[
  scale=0.9,
  anchor=230,
  text=black,
  rotate=0.0
]{ 0.16};
\draw (axis cs:0.8,0.263) node[
  scale=0.9,
  anchor=230,
  text=black,
  rotate=0.0
]{ 0.25};
\draw (axis cs:1.8,0.335) node[
  scale=0.9,
  anchor=230,
  text=black,
  rotate=0.0
]{ 0.32};
\draw (axis cs:2.8,0.38) node[
  scale=0.9,
  anchor=230,
  text=black,
  rotate=0.0
]{ 0.37};
\draw (axis cs:3.8,0.398) node[
  scale=0.9,
  anchor=230,
  text=black,
  rotate=0.0
]{ 0.39};

\nextgroupplot[
axis line style={lightgray204},
enlargelimits=false,
height=5cm,
tick align=outside,
title style={yshift=-1.8ex},
title={Big Bird Embeddings},
unbounded coords=jump,
width=5cm,
x grid style={lightgray204},
xlabel=\textcolor{darkslategray38}{},
xmajorticks=true,
xtick pos=bottom,
xmin=-0.5, xmax=4.5,
xtick style={color=lightgray204},
xtick={0,1,2,3,4},
xticklabels={10,20,40,80,100},
y grid style={lightgray204},
ymajorgrids,
ymajorticks=false,
ymin=0, ymax=1,
ytick style={color=lightgray204}
]
\draw[draw=white,fill=goldenrod22320431] (axis cs:-0.4,0) rectangle (axis cs:0.4,0.262);
\draw[draw=white,fill=goldenrod22320431] (axis cs:0.6,0) rectangle (axis cs:1.4,0.364);
\draw[draw=white,fill=goldenrod22320431] (axis cs:1.6,0) rectangle (axis cs:2.4,0.496);
\draw[draw=white,fill=goldenrod22320431] (axis cs:2.6,0) rectangle (axis cs:3.4,0.622);
\draw[draw=white,fill=goldenrod22320431] (axis cs:3.6,0) rectangle (axis cs:4.4,0.649);
\addplot [line width=1.08pt, darkslategray66]
table {%
0 0
0 0
};
\addplot [line width=1.08pt, darkslategray66]
table {%
1 0
1 0
};
\addplot [line width=1.08pt, darkslategray66]
table {%
2 0
2 0
};
\addplot [line width=1.08pt, darkslategray66]
table {%
3 0
3 0
};
\addplot [line width=1.08pt, darkslategray66]
table {%
4 0
4 0
};
\draw (axis cs:-0.2,0.272) node[
  scale=0.9,
  anchor=230,
  text=black,
  rotate=0.0
]{ 0.26};
\draw (axis cs:0.8,0.374) node[
  scale=0.9,
  anchor=230,
  text=black,
  rotate=0.0
]{ 0.36};
\draw (axis cs:1.8,0.506) node[
  scale=0.9,
  anchor=230,
  text=black,
  rotate=0.0
]{ 0.5};
\draw (axis cs:2.8,0.632) node[
  scale=0.9,
  anchor=230,
  text=black,
  rotate=0.0
]{ 0.62};
\draw (axis cs:3.8,0.659) node[
  scale=0.9,
  anchor=230,
  text=black,
  rotate=0.0
]{ \bfseries 0.65};
\end{groupplot}

\end{tikzpicture}
    \caption{Baseline and embedding model accuracy scores by percentage increments of total input length.}
    \label{fig:text_length}
\end{figure}

\subsection{Error Analysis}
We conducted two additional error analyses to test the robustness of the embeddings: %

\paragraph{Out-of-sample Analysis} \label{sec:out-of-sample}
To assess generalization, we trained 3 classifiers on $D_\text{exp}$ and tested on $D_\text{oos}$. As mentioned, text in $D_\text{oos}$ only concerns European politics, which is close to absent in the training data. In particular, we trained three classifiers using different features: our baseline using the linguistic features, and two classifiers using Big Bird embeddings, using the best performing pre-trained and fine-tuned feature extractors (see Table~\ref{tab:results}). We considered both versions of the feature extractor to control for any data leakage that occurred during fine-tuning.

\paragraph{Sensitivity to Text Length}
To gauge the effect of text length on performance, we randomly sampled 1000 chunks from $D_\text{exp}$ and created slices\footnote{Sliced on \texttt{\textbackslash n}. We also experimented with sentence, clause, and character-level, but observed similar results in all cases.} of 10\%, 20\%, 40\%, and 80\% of the total length of the chunk, following a similar baseline and embedding extraction method as the out-of-sample analysis. Next, we trained a logistic regression classifier, similar to those described in Section \ref{sec:classifiers}, on all of $D_\text{exp}$ except the 1000 randomly sampled chunks. Then, we obtained predictions for all slices, and computed the accuracy for each slice group; i.e., accuracy for all 10\% slices, 20\% slices, etc.

\section{Results} \label{sec:results}

\subsection{Main Experiment}

Table~\ref{tab:results} shows the average CV scores of each classifier. $BigBird_{2048\_tuned}$ yielded the highest average CV accuracy with 65.38\%; a 17 point increase over the baseline trained on linguistic features (47.55\%). The classifiers trained on fine-tuned embeddings outperformed their pre-trained versions across all three model variants. However, differences are smallest for $BigBird_{512}$, suggesting that the short input size limits fine-tuning's efficiency. Increasing input size seems to have a small effect, though we note that the average chunk length in $D_\text{exp}$ is 1726 tokens; i.e., with an input size of 2048 tokens, most are captured already.

\subsection{Embedding Space Analysis}

\begin{figure}[t]
    \centering
    \begin{tikzpicture}

\definecolor{crimson2143940}{RGB}{214,39,40}
\definecolor{darkorange25512714}{RGB}{255,127,14}
\definecolor{darkslategray38}{RGB}{38,38,38}
\definecolor{forestgreen4416044}{RGB}{44,160,44}
\definecolor{lightgray204}{RGB}{204,204,204}
\definecolor{mediumpurple148103189}{RGB}{148,103,189}
\definecolor{sienna1408675}{RGB}{140,86,75}
\definecolor{steelblue31119180}{RGB}{31,119,180}

\begin{groupplot}[group style={group size=2 by 1, horizontal sep=2.2cm}]
\nextgroupplot[
axis line style={lightgray204},
enlargelimits=false,
height=9cm,
tick align=outside,
title style={yshift=-1.4ex},
title={Pretrained Model},
width=4.3cm,
x dir=reverse,
x grid style={lightgray204},
xmajorticks=false,
xmin=0, xmax=1.28467296481331,
xtick style={color=darkslategray38},
y grid style={lightgray204},
ymajorticks=true,
ytick pos=right,
ymin=0, ymax=230,
ytick style={color=darkslategray38},
ytick={5,15,25,35,45,55,65,75,85,95,105,115,125,135,145,155,165,175,185,195,205,215,225},
yticklabel style={anchor=west},
yticklabels={
  Serbian,
  Croatian,
  Romanian,
  Spanish,
  Portuguese,
  Italian,
  French,
  Turkish,
  Russian,
  Slovenian,
  Czech,
  Estonian,
  Hungarian,
  German,
  Polish,
  Lithuanian,
  Greek,
  Bulgarian,
  Norwegian,
  Dutch,
  Finnish,
  Swedish,
  English
}
]
\path [draw=darkorange25512714, semithick]
(axis cs:0,5)
--(axis cs:0.222877217669785,5)
--(axis cs:0.222877217669785,15)
--(axis cs:0,15);

\path [draw=darkorange25512714, semithick]
(axis cs:0.222877217669785,10)
--(axis cs:0.418531153308582,10)
--(axis cs:0.418531153308582,25)
--(axis cs:0,25);

\path [draw=darkorange25512714, semithick]
(axis cs:0,35)
--(axis cs:0.289260813450611,35)
--(axis cs:0.289260813450611,45)
--(axis cs:0,45);

\path [draw=darkorange25512714, semithick]
(axis cs:0.418531153308582,17.5)
--(axis cs:0.496470565212455,17.5)
--(axis cs:0.496470565212455,40)
--(axis cs:0.289260813450611,40);

\path [draw=darkorange25512714, semithick]
(axis cs:0,55)
--(axis cs:0.414225173025582,55)
--(axis cs:0.414225173025582,65)
--(axis cs:0,65);

\path [draw=darkorange25512714, semithick]
(axis cs:0.496470565212455,28.75)
--(axis cs:0.602570522290264,28.75)
--(axis cs:0.602570522290264,60)
--(axis cs:0.414225173025582,60);

\path [draw=darkorange25512714, semithick]
(axis cs:0,75)
--(axis cs:0.601155973930709,75)
--(axis cs:0.601155973930709,85)
--(axis cs:0,85);

\path [draw=darkorange25512714, semithick]
(axis cs:0.602570522290264,44.375)
--(axis cs:0.850258539480154,44.375)
--(axis cs:0.850258539480154,80)
--(axis cs:0.601155973930709,80);

\path [draw=forestgreen4416044, semithick]
(axis cs:0,95)
--(axis cs:0.228559590404589,95)
--(axis cs:0.228559590404589,105)
--(axis cs:0,105);

\path [draw=forestgreen4416044, semithick]
(axis cs:0.228559590404589,100)
--(axis cs:0.250226560048002,100)
--(axis cs:0.250226560048002,115)
--(axis cs:0,115);

\path [draw=forestgreen4416044, semithick]
(axis cs:0.250226560048002,107.5)
--(axis cs:0.29364424014221,107.5)
--(axis cs:0.29364424014221,125)
--(axis cs:0,125);

\path [draw=forestgreen4416044, semithick]
(axis cs:0.29364424014221,116.25)
--(axis cs:0.341783305742762,116.25)
--(axis cs:0.341783305742762,135)
--(axis cs:0,135);

\path [draw=forestgreen4416044, semithick]
(axis cs:0,145)
--(axis cs:0.245914728160391,145)
--(axis cs:0.245914728160391,155)
--(axis cs:0,155);

\path [draw=forestgreen4416044, semithick]
(axis cs:0.341783305742762,125.625)
--(axis cs:0.39540537551966,125.625)
--(axis cs:0.39540537551966,150)
--(axis cs:0.245914728160391,150);

\path [draw=forestgreen4416044, semithick]
(axis cs:0,165)
--(axis cs:0.292039277910627,165)
--(axis cs:0.292039277910627,175)
--(axis cs:0,175);

\path [draw=forestgreen4416044, semithick]
(axis cs:0.39540537551966,137.8125)
--(axis cs:0.477407212595979,137.8125)
--(axis cs:0.477407212595979,170)
--(axis cs:0.292039277910627,170);

\path [draw=forestgreen4416044, semithick]
(axis cs:0,185)
--(axis cs:0.255377364330278,185)
--(axis cs:0.255377364330278,195)
--(axis cs:0,195);

\path [draw=forestgreen4416044, semithick]
(axis cs:0.255377364330278,190)
--(axis cs:0.316180762966012,190)
--(axis cs:0.316180762966012,205)
--(axis cs:0,205);

\path [draw=forestgreen4416044, semithick]
(axis cs:0.316180762966012,197.5)
--(axis cs:0.378542497357674,197.5)
--(axis cs:0.378542497357674,215)
--(axis cs:0,215);

\path [draw=forestgreen4416044, semithick]
(axis cs:0.477407212595979,153.90625)
--(axis cs:0.683244884949426,153.90625)
--(axis cs:0.683244884949426,206.25)
--(axis cs:0.378542497357674,206.25);

\path [draw=steelblue31119180, semithick]
(axis cs:0.850258539480154,62.1875)
--(axis cs:1.04401502304116,62.1875)
--(axis cs:1.04401502304116,180.078125)
--(axis cs:0.683244884949426,180.078125);

\path [draw=steelblue31119180, semithick]
(axis cs:1.04401502304116,121.1328125)
--(axis cs:1.22349806172697,121.1328125)
--(axis cs:1.22349806172697,225)
--(axis cs:0,225);

\nextgroupplot[
axis line style={lightgray204},
enlargelimits=false,
height=9cm,
tick align=outside,
title style={yshift=-1.4ex},
title={Fine-Tuned Model},
width=4.3cm,
x dir=reverse,
x grid style={lightgray204},
xmajorticks=false,
xmin=0, xmax=20.9007087065478,
xtick style={color=darkslategray38},
y grid style={lightgray204},
ymajorticks=true,
ytick pos=right,
ymin=0, ymax=230,
ytick style={color=darkslategray38},
ytick={5,15,25,35,45,55,65,75,85,95,105,115,125,135,145,155,165,175,185,195,205,215,225},
yticklabel style={anchor=west},
yticklabels={
  Swedish,
  Finnish,
  Estonian,
  Bulgarian,
  Norwegian,
  Spanish,
  English,
  Dutch,
  German,
  Portuguese,
  French,
  Greek,
  Turkish,
  Serbian,
  Croatian,
  Slovenian,
  Italian,
  Romanian,
  Hungarian,
  Lithuanian,
  Czech,
  Polish,
  Russian
}
]
\path [draw=darkorange25512714, semithick]
(axis cs:0,5)
--(axis cs:11.4369442453189,5)
--(axis cs:11.4369442453189,15)
--(axis cs:0,15);

\path [draw=darkorange25512714, semithick]
(axis cs:0,25)
--(axis cs:10.6703728828318,25)
--(axis cs:10.6703728828318,35)
--(axis cs:0,35);

\path [draw=darkorange25512714, semithick]
(axis cs:11.4369442453189,10)
--(axis cs:12.1191315047424,10)
--(axis cs:12.1191315047424,30)
--(axis cs:10.6703728828318,30);

\path [draw=forestgreen4416044, semithick]
(axis cs:0,55)
--(axis cs:11.459809283901,55)
--(axis cs:11.459809283901,65)
--(axis cs:0,65);

\path [draw=forestgreen4416044, semithick]
(axis cs:0,75)
--(axis cs:11.1772343754789,75)
--(axis cs:11.1772343754789,85)
--(axis cs:0,85);

\path [draw=forestgreen4416044, semithick]
(axis cs:11.459809283901,60)
--(axis cs:13.3738146037742,60)
--(axis cs:13.3738146037742,80)
--(axis cs:11.1772343754789,80);

\path [draw=crimson2143940, semithick]
(axis cs:0,95)
--(axis cs:11.7434893076754,95)
--(axis cs:11.7434893076754,105)
--(axis cs:0,105);

\path [draw=crimson2143940, semithick]
(axis cs:11.7434893076754,100)
--(axis cs:12.7714543527118,100)
--(axis cs:12.7714543527118,115)
--(axis cs:0,115);

\path [draw=mediumpurple148103189, semithick]
(axis cs:0,135)
--(axis cs:9.03264922983293,135)
--(axis cs:9.03264922983293,145)
--(axis cs:0,145);

\path [draw=mediumpurple148103189, semithick]
(axis cs:9.03264922983293,140)
--(axis cs:10.2344152309427,140)
--(axis cs:10.2344152309427,155)
--(axis cs:0,155);

\path [draw=mediumpurple148103189, semithick]
(axis cs:10.2344152309427,147.5)
--(axis cs:13.8785552940591,147.5)
--(axis cs:13.8785552940591,165)
--(axis cs:0,165);

\path [draw=mediumpurple148103189, semithick]
(axis cs:0,175)
--(axis cs:11.4221096308037,175)
--(axis cs:11.4221096308037,185)
--(axis cs:0,185);

\path [draw=mediumpurple148103189, semithick]
(axis cs:13.8785552940591,156.25)
--(axis cs:13.924640817295,156.25)
--(axis cs:13.924640817295,180)
--(axis cs:11.4221096308037,180);

\path [draw=sienna1408675, semithick]
(axis cs:0,195)
--(axis cs:10.0531031316948,195)
--(axis cs:10.0531031316948,205)
--(axis cs:0,205);

\path [draw=sienna1408675, semithick]
(axis cs:10.0531031316948,200)
--(axis cs:10.7700901521327,200)
--(axis cs:10.7700901521327,215)
--(axis cs:0,215);

\path [draw=sienna1408675, semithick]
(axis cs:10.7700901521327,207.5)
--(axis cs:12.2422167403656,207.5)
--(axis cs:12.2422167403656,225)
--(axis cs:0,225);

\path [draw=steelblue31119180, semithick]
(axis cs:12.1191315047424,20)
--(axis cs:13.9848524282293,20)
--(axis cs:13.9848524282293,45)
--(axis cs:0,45);

\path [draw=steelblue31119180, semithick]
(axis cs:13.9848524282293,32.5)
--(axis cs:16.0836532380491,32.5)
--(axis cs:16.0836532380491,70)
--(axis cs:13.3738146037742,70);

\path [draw=steelblue31119180, semithick]
(axis cs:12.7714543527118,107.5)
--(axis cs:14.8844749285999,107.5)
--(axis cs:14.8844749285999,125)
--(axis cs:0,125);

\path [draw=steelblue31119180, semithick]
(axis cs:16.0836532380491,51.25)
--(axis cs:17.6498989836203,51.25)
--(axis cs:17.6498989836203,116.25)
--(axis cs:14.8844749285999,116.25);

\path [draw=steelblue31119180, semithick]
(axis cs:13.924640817295,168.125)
--(axis cs:17.0276693723796,168.125)
--(axis cs:17.0276693723796,216.25)
--(axis cs:12.2422167403656,216.25);

\path [draw=steelblue31119180, semithick]
(axis cs:17.6498989836203,83.75)
--(axis cs:19.9054368633789,83.75)
--(axis cs:19.9054368633789,192.1875)
--(axis cs:17.0276693723796,192.1875);

\end{groupplot}

\end{tikzpicture}
    \caption{Hierarchical clustering dendograms of native language centroids in the 768-dimensional Big Bird embedding space before and after fine-tuning.}
    \label{fig:language_clustering}
\end{figure}
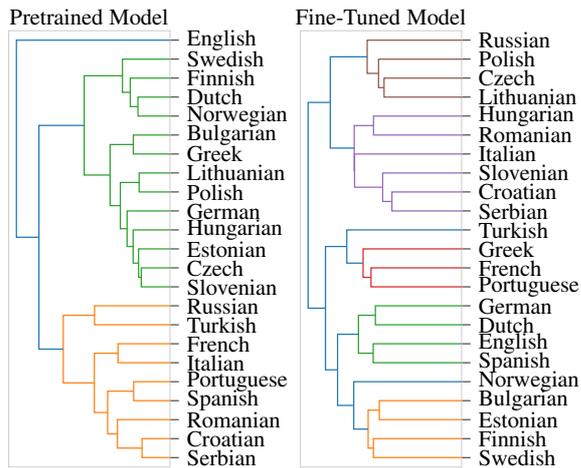

\begin{figure}[t]
    \centering
    \definecolor{darkslategray38}{RGB}{38,38,38}
\definecolor{darkslategray66}{RGB}{66,66,66}
\definecolor{goldenrod22320431}{RGB}{223,204,31}
\definecolor{lightgray204}{RGB}{204,204,204}
\definecolor{peru22412844}{RGB}{224,128,44}

\begin{tikzpicture}

\definecolor{darkslategray38}{RGB}{38,38,38}
\definecolor{lightgray204}{RGB}{204,204,204}
\definecolor{steelblue31119180}{RGB}{31,119,180}

\begin{groupplot}[group style={group size=2 by 1, horizontal sep=0.2cm}]
\nextgroupplot[
axis line style={lightgray204},
enlargelimits=false,
height=9cm,
tick align=outside,
title style={yshift=-1.4ex},
title={Pretrained Model},
width=6cm,
x grid style={lightgray204},
xlabel=\textcolor{darkslategray38}{},
xmajorticks=false,
xmin=-0.345002016425133, xmax=0.17488451898098,
xtick style={color=darkslategray38},
y grid style={lightgray204},
ylabel=\textcolor{darkslategray38}{},
ymajorticks=false,
ymin=-0.226713313907385, ymax=0.44192358776927,
ytick style={color=darkslategray38}
]
\addplot [draw=steelblue31119180, fill=steelblue31119180, mark=*, only marks]
table{%
x  y
-0.321370810270309 0.411531001329422
0.108572989702225 0.117562368512154
-0.0270839389413595 0.0959853604435921
-0.0337204337120056 -0.0888960212469101
0.0107075739651918 -0.0103378528729081
0.0526995472609997 0.0433132313191891
-0.132168024778366 0.136405184864998
-0.113247662782669 -0.0411282628774643
-0.0758965238928795 0.000550828000996262
0.0544491112232208 -0.0231173411011696
-0.0851535722613335 -0.196320727467537
0.1340072453022 -0.00257052527740598
-0.0443454273045063 0.131003856658936
-0.00489763310179114 0.1563750654459
0.125616654753685 -0.079339787364006
0.0182028263807297 -0.188316702842712
0.14825913310051 -0.176354140043259
-0.0104159666225314 -0.0827919021248817
0.0890020579099655 -0.117051869630814
0.0326680913567543 0.0766520425677299
0.0275096483528614 -0.106718152761459
0.151253312826157 -0.0189445409923792
-0.0353056266903877 -0.144636884331703
};
\draw (axis cs:-0.24370810270309,0.396531001329422) node[
  scale=1.0,
  text=darkslategray38,
  rotate=0.0
]{English};
\draw (axis cs:0.0935729897022247,0.148562368512154) node[
  scale=1.0,
  text=darkslategray38,
  rotate=0.0
]{German};
\draw (axis cs:-0.1220839389413595,0.0809853604435921) node[
  scale=1.0,
  text=darkslategray38,
  rotate=0.0
]{Bulgarian};
\draw (axis cs:-0.1187204337120056,-0.08389602124691) node[
  scale=1.0,
  text=darkslategray38,
  rotate=0.0
]{Croatian};
\draw (axis cs:-0.04429242603480816,-0.0253378528729081) node[
  scale=1.0,
  text=darkslategray38,
  rotate=0.0
]{Czech};
\draw (axis cs:-0.0326995472609997,0.0463132313191891) node[
  scale=1.0,
  text=darkslategray38,
  rotate=0.0
]{Estonian};
\draw (axis cs:-0.207168024778366,0.151405184864998) node[
  scale=1.0,
  text=darkslategray38,
  rotate=0.0
]{Finnish};
\draw (axis cs:-0.19247662782669,-0.0301282628774643) node[
  scale=1.0,
  text=darkslategray38,
  rotate=0.0
]{French};
\draw (axis cs:-0.1408965238928795,0.0044491719990037) node[
  scale=1.0,
  text=darkslategray38,
  rotate=0.0
]{Greek};
\draw (axis cs:-0.0094491112232208,-0.0481173411011696) node[
  scale=1.0,
  text=darkslategray38,
  rotate=0.0
]{Hungarian};
\draw (axis cs:-0.150153572261333,-0.211320727467537) node[
  scale=1.0,
  text=darkslategray38,
  rotate=0.0
]{Italian};
\draw (axis cs:0.0790072453022,0.016429474722594) node[
  scale=1.0,
  text=darkslategray38,
  rotate=0.0
]{Lithuanian};
\draw (axis cs:-0.0893454273045063,0.110003856658936) node[
  scale=1.0,
  text=darkslategray38,
  rotate=0.0
]{Dutch};
\draw (axis cs:0.0151023668982089,0.1813750654459) node[
  scale=1.0,
  text=darkslategray38,
  rotate=0.0
]{Norwegian};
\draw (axis cs:0.110616654753685,-0.094339787364006) node[
  scale=1.0,
  text=darkslategray38,
  rotate=0.0
]{Polish};
\draw (axis cs:0.06320282638072968,-0.213316702842712) node[
  scale=1.0,
  text=darkslategray38,
  rotate=0.0
]{Portuguese};
\draw (axis cs:0.08025913310051,-0.161354140043259) node[
  scale=1.0,
  text=darkslategray38,
  rotate=0.0
]{Romanian};
\draw (axis cs:0.00458403337746859,-0.0677919021248817) node[
  scale=1.0,
  text=darkslategray38,
  rotate=0.0
]{Russian};
\draw (axis cs:0.104002057909966,-0.132051869630814) node[
  scale=1.0,
  text=darkslategray38,
  rotate=0.0
]{Serbian};
\draw (axis cs:0.0876680913567543,0.10165204256773) node[
  scale=1.0,
  text=darkslategray38,
  rotate=0.0
]{Slovenian};
\draw (axis cs:-0.0425096483528614,-0.121718152761459) node[
  scale=1.0,
  text=darkslategray38,
  rotate=0.0
]{Spanish};
\draw (axis cs:0.131253312826157,-0.0339445409923792) node[
  scale=1.0,
  text=darkslategray38,
  rotate=0.0
]{Swedish};
\draw (axis cs:-0.1003056266903877,-0.159636884331703) node[
  scale=1.0,
  text=darkslategray38,
  rotate=0.0
]{Turkish};

\nextgroupplot[
axis line style={lightgray204},
enlargelimits=false,
height=9cm,
tick align=outside,
title style={yshift=-1.4ex},
title={Fine-Tuned Model},
width=6cm,
x grid style={lightgray204},
xlabel=\textcolor{darkslategray38}{},
xmajorticks=false,
xmin=-5.57091040611267, xmax=6.61993975639343,
xtick style={color=darkslategray38},
y grid style={lightgray204},
ymajorticks=false,
ymin=-5.53120133876801, ymax=3.88225886821747,
ytick style={color=darkslategray38}
]
\addplot [draw=steelblue31119180, fill=steelblue31119180, mark=*, only marks]
table{%
x  y
-4.17060708999634 -2.20591974258423
-2.85464429855347 1.27809429168701
-1.38378155231476 1.80048501491547
1.76037204265594 0.847255051136017
5.8740930557251 1.65857899188995
1.52379834651947 3.23564028739929
1.97196006774902 3.0368926525116
-1.73588275909424 -5.10331678390503
-1.88456916809082 0.0307963788509369
0.673154830932617 1.40693747997284
-1.3540370464325 -2.68723583221436
6.06581020355225 2.20650696754456
-5.01678085327148 1.88221800327301
-2.65406656265259 0.0354563407599926
5.73623752593994 0.560691475868225
-2.49010252952576 -2.09098100662231
0.138834878802299 0.528136372566223
3.73294258117676 -2.54856705665588
2.4094021320343 -0.78379362821579
1.83526682853699 3.45437431335449
-3.20129704475403 -4.58792161941528
-2.54192638397217 1.93159806728363
1.02461767196655 -3.06198716163635
};
\draw (axis cs:-3.5060708999634,-2.60591974258423) node[
  scale=1.0,
  text=darkslategray38,
  rotate=0.0
]{English};
\draw (axis cs:-3.70464429855347,1.07809429168701) node[
  scale=1.0,
  text=darkslategray38,
  rotate=0.0
]{German};
\draw (axis cs:0.88378155231476,1.75048501491547) node[
  scale=1.0,
  text=darkslategray38,
  rotate=0.0
]{Bulgarian};
\draw (axis cs:3.7037204265594,0.847255051136017) node[
  scale=1.0,
  text=darkslategray38,
  rotate=0.0
]{Croatian};
\draw (axis cs:4.2740930557251,1.45857899188995) node[
  scale=1.0,
  text=darkslategray38,
  rotate=0.0
]{Czech};
\draw (axis cs:-0.70379834651947,3.19564028739929) node[
  scale=1.0,
  text=darkslategray38,
  rotate=0.0
]{Estonian};
\draw (axis cs:3.807196006774902,3.0368926525116) node[
  scale=1.0,
  text=darkslategray38,
  rotate=0.0
]{Finnish};
\draw (axis cs:0.03588275909424,-5.30331678390503) node[
  scale=1.0,
  text=darkslategray38,
  rotate=0.0
]{French};
\draw (axis cs:-0.28456916809082,-0.169203621149063) node[
  scale=1.0,
  text=darkslategray38,
  rotate=0.0
]{Greek};
\draw (axis cs:0.473154830932617,1.20693747997284) node[
  scale=1.0,
  text=darkslategray38,
  rotate=0.0
]{Hungarian};
\draw (axis cs:0.1040370464325,-2.48723583221436) node[
  scale=1.0,
  text=darkslategray38,
  rotate=0.0
]{Italian};
\draw (axis cs:3.56581020355225,2.20650696754456) node[
  scale=1.0,
  text=darkslategray38,
  rotate=0.0
]{Lithuanian};
\draw (axis cs:-3.51678085327148,1.68221800327301) node[
  scale=1.0,
  text=darkslategray38,
  rotate=0.0
]{Dutch};
\draw (axis cs:-3.4406656265259,0.364543659240007) node[
  scale=1.0,
  text=darkslategray38,
  rotate=0.0
]{Norwegian};
\draw (axis cs:4.53623752593994,0.360691475868225) node[
  scale=1.0,
  text=darkslategray38,
  rotate=0.0
]{Polish};
\draw (axis cs:-0.49010252952576,-1.84098100662231) node[
  scale=1.0,
  text=darkslategray38,
  rotate=0.0
]{Portuguese};
\draw (axis cs:1.0611651211977005,0.328136372566223) node[
  scale=1.0,
  text=darkslategray38,
  rotate=0.0
]{Romanian};
\draw (axis cs:4.53294258117676,-2.94856705665588) node[
  scale=1.0,
  text=darkslategray38,
  rotate=0.0
]{Russian};
\draw (axis cs:4.1094021320343,-0.98379362821579) node[
  scale=1.0,
  text=darkslategray38,
  rotate=0.0
]{Serbian};
\draw (axis cs:4.08526682853699,3.654374331335449) node[
  scale=1.0,
  text=darkslategray38,
  rotate=0.0
]{Slovenian};
\draw (axis cs:-1.20129704475403,-4.3792161941528) node[
  scale=1.0,
  text=darkslategray38,
  rotate=0.0
]{Spanish};
\draw (axis cs:-2.74192638397217,2.23159806728363) node[
  scale=1.0,
  text=darkslategray38,
  rotate=0.0
]{Swedish};
\draw (axis cs:0.824617671966553,-3.36198716163635) node[
  scale=1.0,
  text=darkslategray38,
  rotate=0.0
]{Turkish};
\end{groupplot}

\end{tikzpicture}
    \caption{2-dimensional PCA space showing the language centroids before and after fine-tuning.}
    \label{fig:pca}
\end{figure}
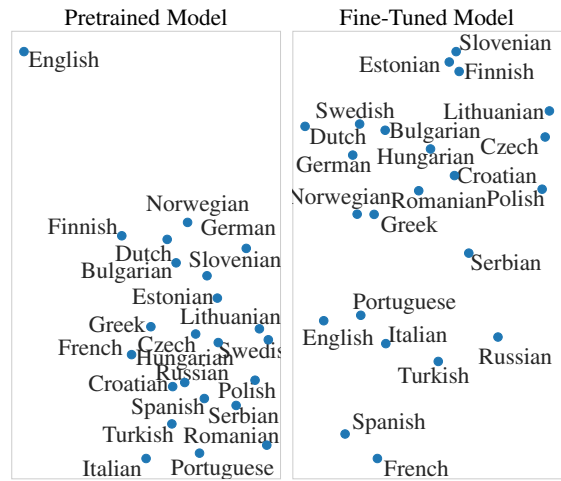

Although our clustering shows some overlap with the results of \newcite{rabinovich-2018-reddit-l2}, there are some deviations. Languages from the same language family are not always close (see Figure~\ref{fig:language_clustering}, fine-tuned or not). For example, Russian is clustered with Turkish (pre-trained) and Italian with the former Yugoslavian languages (fine-tuned). 
Furthermore, fine-tuning shifts the embedding space more toward separating individual languages, rather than separating native-English from non-native English (as indicated by English having it own cluster).   %
This effect is most apparent in the low-dimensional PCA space (see Figure~\ref{fig:pca}). %
In the fine-tuned space, an interesting artifact can be observed, where the space roughly mimics the languages' geographical orientation to each other.

\subsection{Error Analysis}
\paragraph{Out-of-sample Analysis}
Here we see the same pattern as in our main experiment (see Table~\ref{tab:results}), with the fine-tuned embedding approach yielding the most accurate classifier, outperforming the feature engineering baseline by 22 percentage points, whereas the pre-trained model gains 13.7.

\paragraph{Sensitivity to Text Length}

In Figure \ref{fig:text_length}, it can be observed that the performance of both embedding and feature engineering classifiers deteriorates as text length decreases. However, the deterioration is not linear, which suggests  %
there is increased redundancy in the information used for classification the longer the input becomes. 
The embeddings are more affected, with a 12 point drop when reducing from 80\% to 40\% and a 14 point drop when reducing from 40\% to 20\%, compared to 5 points and 7 points for the feature engineering model. %

\section{Discussion \& Conclusion}

Our experiments demonstrate how fairly straightforward featurization using embeddings from transformers that account for long enough input sequences is faster, and substantially outperforms prior best performing models. Some limitations should be mentioned, such as the restricted domain (Reddit only), the dataset containing mostly highly fluent English speakers, and English being the only L2. Moreover, while out-of-sample, $D_\text{oos}$ was likely not completely new; Big Bird might have been trained on Reddit prior, and, therefore, other social platforms are worth evaluating on as well (although label collection will likely be significantly more challenging). We expect even better results if other classifiers are used and tuned, and a comparison with similar transformers such as Longformer \cite{longformer-paper} and Transformer-XL \cite{transformer-xl-paper} is certainly worthwhile \cite{scaling-to-1m-tokens}. As is commonly observed \cite{bert-original-paper-2018,chi-finetuning-bert-2019,universal-fine-tuning}, fine-tuning Big Bird on our data improved performance, and our observations proved robust both throughout cross-validation and on out-of-sample data. Given these results, we believe our works offers a promising avenue for future NLI work.

\section{Acknowledgments}

Our research strongly relied on openly available resources.  We thank all whose work we could use.

\bibliographystyle{acl_natbib}
\bibliography{references}

\end{document}